# Dynamic Boltzmann Machines for Second Order Moments and Generalized Gaussian Distributions

**Rudy Raymond**
IBM Research - Tokyo
rudyhar@jp.ibm.com

**Takayuki Osogami**
IBM Research - Tokyo
osogami@jp.ibm.com

**Sakyasingha Dasgupta**
LeapMind, Inc.*
sakya@leapmind.io

## Abstract

Dynamic Boltzmann Machine (DyBM) has been shown highly efficient to predict time-series data. Gaussian DyBM is a DyBM that assumes the predicted data is generated by a Gaussian distribution whose first-order moment (mean) dynamically changes over time but its second-order moment (variance) is fixed. However, in many financial applications, the assumption is quite limiting in two aspects. First, even when the data follows a Gaussian distribution, its variance may change over time. Such variance is also related to important temporal economic indicators such as the market volatility. Second, financial time-series data often requires learning datasets generated by the generalized Gaussian distribution with an additional shape parameter that is important to approximate heavy-tailed distributions. Addressing those aspects, we show how to extend DyBM that results in significant performance improvement in predicting financial time-series data.

## 1 Introduction

DyBM is an artificial model of a spiking neural network [13, 11] and has been modified to deal with real-valued time-series in the form of Gaussian DyBM (G-DyBM) [10, 4]. G-DyBM has been shown effective to predict time-varying real values on the basis of the assumption that the underlying distribution is Gaussian whose variance is fixed but its mean changes over time. In particular, G-DyBM is known to be related to the vector autoregressive (VAR) model [9]: a special case of the G-DyBM is a VAR model with additional variables capturing long-term dependencies [10]. In case of binary time-series, the additional variables correspond to the timing and frequencies of spikes arriving from one neuron (or, unit) to another, which is shown to be related to the spike-timing dependent plasticity (STDP). The STDP to the DyBM is akin to the Hebb's rule to the Boltzmann Machine [1].

Experiments have confirmed that G-DyBM improved the VAR models significantly. Furthermore, it can easily be extended to more complex models incorporating non-linearity at a fraction of computational cost of the long short-term memory (LSTM) networks [4]. Recently, G-DyBM has also been extended with hidden units [12] and with functional time-series prediction [8], which essentially extends VAR and functional autoregression [3]. Nevertheless, the G-DyBM models assume Gaussian distributions with fixed variances and hence cannot be readily used to model financial time series data whose underlying distributions are often heavy-tailed and whose variances change over time. Generalized Autoregressive Conditional Heteroskedasticity (GARCH) models are widely popular for predicting time-dependent variances since the seminal paper by Engle [6].

We first show how to extend G-DyBM to predict the time-varying variances. In this case, the extended G-DyBM models are comparable to the GARCH models, just as the G-DyBM models naturally extend the VAR models. We show that, similar to the GARCH$(1, 1)$ model that gives a closed-form equation of n-step ahead of variance prediction, the extended G-DyBM can yield a similar, albeit complex, closed-form equation. We further show an extension of G-DyBM to learn the generalized Gaussian distribution. We empirically confirm that the extended G-DyBM models consistently improve their corresponding baseline methods on real-world datasets.

---

*This work was done while the author was at IBM Research - Tokyo.

## 2 Gaussian DyBM for predicting real-valued time series

G-DyBM predicts the next real-valued time series by incorporating information from past sequence of the time-series. G-DyBM uses First-In-First-Out (FIFO) queues, to store the most recent sequences of time-series, and eligibility traces, to store the summary of past time-series.

Letting the multi-dimensional time-series sequence be $\{\mathbf{x}^{[1]}, \mathbf{x}^{[2]}, \ldots, \mathbf{x}^{[t-1]}\}$, G-DyBM predicts the values at time $t$ by the following equation:

$$\boldsymbol{\mu}^{[t]} = \mathbf{b} + \sum_{\delta=1}^{d-1} \mathbf{W}^{[\delta]} \mathbf{x}^{[t-\delta]} + \sum_{k=1}^{K} \mathbf{U}^{[k]} \boldsymbol{\alpha}_k^{[t-\delta]}, \quad (1)$$

where $\mathbf{b}, \mathbf{W}^{[\delta]}, \mathbf{U}^{[k]}$ are internal parameters of G-DyBM, and $\boldsymbol{\alpha}_k^{[t-\delta]} = \sum_{s=-\infty}^{t-d} \lambda_k^{t-s-d} \mathbf{x}^{[s]}$ is the vector of eligibility traces with decaying factors $0 < \lambda_k < 1$, which are assumed given and fixed.

G-DyBM is comparable to the standard VAR model. Specifically, the second term on the right-hand side of the above equation is the $d-1$ lag of the VAR model, while the last term, unique to G-DyBM, corresponds to the eligibility traces summarizing the entire past sequences.

Another important aspect that differentiates G-DyBM from VAR is that it learns and adjusts the values of its internal parameters by online updates taking into account the discrepancy of its predictions with the true values. The internal parameters of G-DyBM are then updated in an online manner as detailed in [10, 4], which is essentially performed to maximize the loglikelihood under Gaussian distributions. Namely, for each $j$, each element $x_j^{[t]}$ of the $\mathbf{x}^{[t]}$ is assumed to follow a Gaussian distribution:

$$p\left(x_j^{[t]} \mid \mathbf{x}^{[<t]}\right) = \frac{1}{\sqrt{2\pi\sigma_j^2}} \exp\left(-\frac{(x_j^{[t]} - \mu_j^{[t]})^2}{2\sigma_j^2}\right), \quad (2)$$

where $\mu_j^{[t]}$ is given by Eq. (1), and $\sigma_j^2$ is G-DyBM's another internal parameter representing the variance, which is assumed to be fixed for the entire sequence. The Gaussian distribution assumption poses two limitations when dealing with data generated from time-varying variances and heavy-tailed distributions. In the hereafter, we propose novel DyBM models to overcome the limitations.

## 3 DyBM for predicting changing variances

Here, we omit the subscript $j$ and use $x_t$ for $x^{[t]}$ for simplicity. The original G-DyBM assumes fixed standard deviation, namely, $(x_t - \mu_t) = \epsilon_t$, where $\epsilon_t$ follows a Gaussian (or, Normal) distribution with mean 0 and standard deviation $\sigma$, i.e., $\epsilon_t \sim N(0, \sigma)$. Following the GARCH model, we consider the case when $(x_t - \mu_t) = \sigma_t \epsilon_t$, where $\epsilon_t \sim N(0, 1)$. The so-called GARCH$(p, q)$ predicts the next-step standard deviation from the most recent $p$ errors and $q$ predictions, as below:

$$\sigma_t^2 = a_0 + \sum_{i=1}^{p} a_i (x_{t-i} - \mu_t)^2 + \sum_{j=1}^{q} b_j \sigma_{t-j}^2. \quad (3)$$

In particular, the popular GARCH$(1, 1)$ reduces to:

$$\sigma_t^2 = a_0 + a_1 e_{t-1}^2 + b_1 \sigma_{t-1}^2 = \frac{a_0}{1 - b_1} + a_1 e_{t-1}^2 + a_1 \sum_{i=1}^{\infty} b_1^i e_{t-1-i}^2 \quad (4)$$

for $|b_1| < 1$, where $e_t \equiv (x_t - \mu_t)$. To impose that the prediction values stay nonnegative and converge, in the above $a_0, a_1, b_1 \geq 0$ and $a_1 + b_1 < 1$ are assumed [14].

On the other hand, with regards to GARCH$(p, q)$, we can consider G-DyBM$(d, k)$, that takes into account $d$ lags and $k$ modes of eligibility traces, as below:

$$\sigma_t^2 = v_0 + \sum_{i=1}^{d} w_i e_{t-i}^2 + \sum_{j=1}^{k} u_j \sum_{i=d}^{\infty} \lambda_k^{i-d+1} e_{t-1-i}^2. \quad (5)$$

In particular, G-DyBM(1, 1) that corresponds to GARCH(1, 1) reduces to:

$$\sigma_t^2 = v_0 + w_1 e_{t-1}^2 + u_1 \sum_{i=1}^{\infty} \lambda_1^i e_{t-1-i}^2, \qquad (6)$$

where we can see easily that G-DyBM(1, 1) is equal to GARCH(1, 1) when $v_0 = \frac{a_0}{1-b_1}$, $w_1 = u_1 = a_1$, and $\lambda_1 = b_1$. We can confirm that G-DyBM$(d, k)$ is more general than GARCH$(p, q)$. Due to its simplicity, the parsimonious GARCH(1, 1) is often popular to predict time-varying standard deviations. By some standard algebra transformation, one can show that Eq. (4) can be used to predict the next $n$-steps of variances $\sigma_{t+n}^2$ from sequences up to time $t$, as below.

$$\sigma_{t+n}^2 = \sigma^2 + (a_1 + b_1)^n \left( \sigma_t^2 - \sigma^2 \right), \qquad (7)$$

where $\sigma^2 \equiv \frac{a_0}{1-a_1-b_1}$. Similarly, we can show that G-DyBM(1, 1) can be used to derive a closed-form equation of the next $n$ steps of variances, as follows.

$$\sigma_{t+n}^2 = \alpha + C_0 + C_1 r_1^n + C_2 r_2^n, \qquad (8)$$

where the values of $\alpha, C_0, C_1, r_1, r_2$ as well as the derivation of the above closed-form equation are shown in the Appendix 7.1.

The extended G-DyBM to predict both time-varying means and standard deviations can be straight-forwardly derived as follows. We use the standard G-DyBM to predict the time-varying means and obtain $\boldsymbol{\mu}^{[t]}$. For any $j$, let the prediction error for $e_t \equiv (x_t - \mu_t)$, and then we can use Eq. (5) to predict the variances at the next step.

## 4 DyBM for predicting generalized Gaussian distribution

The $\boldsymbol{\mu}^{[t]}$ in Eq. (1) is the maximum likelihood estimator of $\mathbf{x}^{[t]}$ under the assumption of Eq. (2). In this subsection, we consider the case when $x_j^{[t]}$ follows a generalized Gaussian distribution as follows:

$$p_j \left( x_j^{[t]} | \mathbf{x}^{[t-T, t-1]} \right) = \frac{\beta^{1/2}}{2\Gamma(1 + 1/\rho)} \exp \left( -\beta^{\frac{\rho}{2}} \left| x_j^{[t]} - \mu_j^{[t]} \right|^\rho \right), \qquad (9)$$

where $\rho > 0$ is the shape parameter ($= 2$ if Gaussian), $\beta > 0$ is the inverse variance, and $\Gamma(\cdot)$ is the gamma function. Similar to the G-DyBM, to compute the maximum likelihood we assume that $p_j(\cdot)$ is independent of $p'_j(\cdot)$ for $j \neq j'$. The log-likelihood function for the generalized G-DyBM is therefore can be written as:

$$\text{LL} \equiv \sum_j \frac{1}{2} \ln \beta_j - \ln \Gamma(1 + 1/\rho_j) - \left( \beta_j (x_j^{[t]} - \mu_j^{[t]})^2 \right)^{\rho_j/2}. \qquad (10)$$

From Eq. (10) we can derive the online update rules for the internal parameters of the extended G-DyBM by taking into account the partial derivatives of LL, as below (omitting the scripts):

$$\frac{\partial \text{LL}}{\partial \beta} = \frac{1}{2\beta} - \frac{\rho}{2} \beta^{\frac{\rho}{2}-1} |x - \mu|^\rho \qquad (11)$$

$$\frac{\partial \text{LL}}{\partial \rho} = \frac{1}{\rho^2} \Psi \left( 1 + \frac{1}{\rho} \right) - \frac{(\beta(x-\mu)^2)^{\frac{\rho}{2}} \ln \beta (x-\mu)^2}{2} \qquad (12)$$

$$\frac{\partial \text{LL}}{\partial \mu} = \beta^{\frac{\rho}{2}} \rho \, \text{sign}(x-\mu) |x-\mu|^{\rho-1}, \qquad (13)$$

where $\Psi$ is the digamma function, which is the logarithmic derivative of the gamma function.

Notice that there are more internal parameters, and estimating both the shape and inverse variance parameters is hard. For this, we employ a heuristic of adjusting the online updates after a fixed number of gradient updates. Let $T$ be the current number of updates. Then, we revise the inverse variance $\beta$ by $\beta = \left( \frac{T}{\rho \sum_t |x^{[t]} - \mu^{[t]}|^\rho} \right)^{\frac{2}{\rho}}$. Thereafter, we update the shape parameter $\rho$ by incorporating techniques following [5] as detailed in the Appendix 7.2. The adjustments enable us to guide the parameter updates for better approximations as confirmed in the experiments.

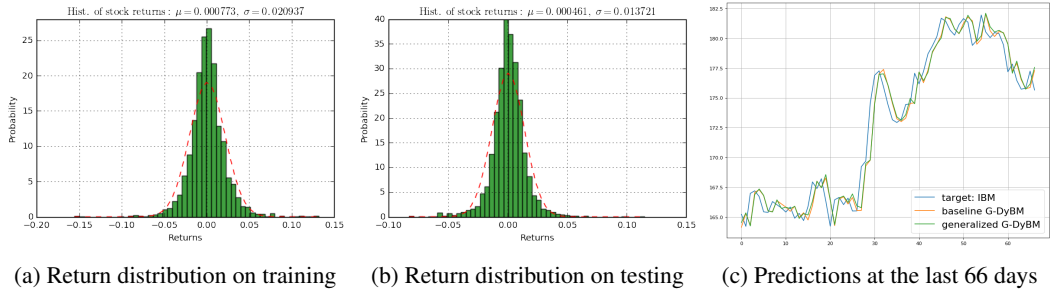

| (a) Return distribution on training | (b) Return distribution on testing | (c) Predictions at the last 66 days |

Figure 1: The first two subfigures show the distributions whose tails are heavier than normal. The last subfigure depicts the target and the predictions of generalized G-DyBM and the baseline G-DyBM.

## 5 Experiments

Here, we evaluate the performances of the proposed DyBM models on the adjusted price of daily closing price of IBM stock from Jan. 3, 1995 to Mar. 17, 2017. The dataset consists of 5592 real-valued time series. As preprocessing, we transform the time series of daily closing stock prices $p_t$ into the time series of daily returns $r_t$, where $r_t \equiv \frac{p_t - p_{t-1}}{p_{t-1}}$. They are then scaled by their standard deviation. Figures (1a) and (1b) show the distribution of returns during the first and second half, where the red dotted curves denote their Gaussian approximations. We can see that the mean and variance of the two distributions are different, and thus justify the online learning with DyBM. More importantly, we can observe that the distributions are asymmetric, and exhibit heavier tails than Gaussian distributions.

We first compare the G-DyBM against the generalized Gaussian DyBM that assumes time series generated as in Eq. (9). Fig. (1c) shows that the generalized G-DyBM can predict better than the baseline G-DyBM. We trained the DyBMs with the first 5526 days of stock prices to predict those of the last 66 days. We obtain that the Root Mean Squared Errors (RMSE, the lower the better) of generalized Gaussian DyBM on the training and testing datasets are $1.509$ and $1.365$, resp. These are lower than the baseline G-DyBM that achieves RMSE of $1.512$ and $1.372$, resp. We run both models to read the time series in 5 epochs with the parameters lag $d = 66$, two decay rates $\lambda_1 = 0.1$, $\lambda_2 = 0.9$, and learning rate $\eta = 0.01$. The generalized G-DyBM adjusts the shape and inverse variance at every $T = 100$ steps. At the end of the testing, it finds the shape parameter $\rho = 0.92$, which is close to a Laplacian distribution.

We then compare the GARCH$(1,1)$ against G-DyBM$(1,1)$ to predict time-varying second moment. In case of GARCH$(1,1)$, we use the *arch* package [2] that provides standard volatility models for Eq. (4). For G-DyBM$(1,1)$, we use the batch version of DyBM package [3] that enables us to obtained positivity constraints by the $L1$ regularization for Eq. (6). We spit the data into two: the first half for training and the rest for testing. We fix the rate $\lambda_1 = 0.97$. The Pearson correlations (the higher the better) of GARCH$(1,1)$ on the training and testing datasets are $0.32$ and $0.39$, resp., Meanwhile, those of G-DyBM$(1,1)$ are higher: they are $0.43$ and $0.41$, resp. We used the time series of $\mu_t$ from the Gaussian DyBM model.

## 6 Conclusion

We proposed new DyBM models for predicting time-series data whose second order moment of the underlying distribution changes over time, and for predicting time-series data with generalized Gaussian distributions. We showed that the former generalizes the popular univariate GARCH models, and both can be used for better prediction of financial datasets. More extensive experiments and extending DyBM to advanced models accommodating asymmetry [7], multivariate GARCH [2], and other important distributions are some of interesting future work.

---

[2] Available from https://pypi.python.org/pypi/arch/4.0

[3] Available from https://github.com/ibm-research-tokyo/dybm

**Acknowledgments**

This work was supported by JST CREST Grant Number JPMJCR1304, Japan.

# 7 Appendix

## 7.1 Predicting $N$-step ahead variances by G-DyBM$(1,1)$

We show the derivation of the closed-form equation shown in Eq. (8) along with the values of $\alpha, C_0, C_1, r_1, r_2$. Notice that by Eq. (6), we have

$$\sigma_t^2 = v_0 + w_1 e_{t-1}^2 + u_1 \sum_{i=2}^{\infty} \lambda_1^{i-1} e_{t-i}^2.$$

Expanding the above equation and arranging the coefficients, we obtain

$$\left(\sigma_{t+N}^2 - \frac{v_0(1-\lambda_1)}{1-w_1-\lambda_1}\right) = (w_1 + \lambda_1)\left(\sigma_{t+N-1}^2 - \frac{v_0(1-\lambda_1)}{1-w_1-\lambda_1}\right) + \lambda_1(u_1 - w_1)\sigma_{t+N-2}^2, \quad (14)$$

for $N = 1, 2, 3, \ldots$. Letting $\alpha \equiv \frac{v_0(1-\lambda_1)}{1-(w_1+\lambda_1)}$, $\beta \equiv (w_1 + \lambda_1)$, and $\gamma \equiv \lambda_1(u_1 - w_1)$, Eq. (14) can be rewritten as below.

$$\left(\sigma_{t+N}^2 - \alpha\right) = \beta\left(\sigma_{t+N-1}^2 - \alpha\right) + \gamma \sigma_{t+N-2}^2, \quad (15)$$

for $N = 1, 2, 3, \ldots$. Defining $S_{t+N} \equiv \left(\sigma_{t+N}^2 - \alpha\right)$, the above Eq. (15) can be rewritten as

$$S_{t+N} - \beta S_{t+N-1} - \gamma S_{t+N-2} = \alpha\gamma, \quad (16)$$

which is easily seen to have the solution in the form of

$$S_{t+N} = C_0 + C_1 r_1^N + C_2 r_2^N, \quad (17)$$

where $C_0 \equiv \frac{\gamma}{1-\beta-\gamma}\alpha$, and $r_1$ and $r_2$ are the solutions to the quadratic equation $r^2 - \beta r - \gamma = 0$, namely, $r_{1,2} = \frac{\beta \pm \sqrt{\beta^2 + 4\gamma}}{2}$. The values of $C_1$ and $C_2$ can be determined from the boundary conditions. Namely, at $N = 0$ we have

$$S_t \equiv \sigma_t^2 - \alpha = C_0 + C_1 + C_2,$$

and at $N = 1$ we have

$$S_{t+1} \equiv \sigma_{t+1}^2 - \alpha = C_0 + C_1 r_1 + C_2 r_2.$$

Finally, we have the closed-form equation of the $N$-step ahead variances from the following values.

$$\alpha \equiv \frac{v_0(1-\lambda_1)}{1-(w_1+\lambda_1)} \quad (18)$$

$$\beta \equiv w_1 + \lambda_1 \quad (19)$$

$$\gamma \equiv \lambda_1(u_1 - w_1) \quad (20)$$

$$r_1 \equiv \frac{\beta + \sqrt{\beta^2 + 4\gamma}}{2} \quad (21)$$

$$r_2 \equiv \frac{\beta - \sqrt{\beta^2 + 4\gamma}}{2} \quad (22)$$

$$C_0 \equiv \frac{\gamma\alpha}{1-\beta-\gamma} \quad (23)$$

$$C_1 \equiv \frac{1}{r_1 - r_2}\left(\gamma e_{t-1}^2 + r_1\left(\sigma_t^2 - \alpha - C_0\right) - C_0(1-\beta)\right) \quad (24)$$

$$C_2 \equiv -\frac{1}{r_1 - r_2}\left(\gamma e_{t-1}^2 + r_2\left(\sigma_t^2 - \alpha - C_0\right) - C_0(1-\beta)\right). \quad (25)$$

## 7.2 Updating shape parameters of Generalized Gaussian Distribution

The shape parameter $\rho$ is updated after every fixed number of online gradient updates according to the following estimation, that is shown in [5].

We first compute the following constant $c$,

$$c = \frac{\left(\frac{1}{T}\sum_t \left|x_j^{[t]} - \mu_j^{[t]}\right|\right)^2}{\frac{1}{T}\sum_t \left|x_j^{[t]} - \mu_j^{[t]}\right|^2}.$$

With regards to the above $c$, we then compute the new shape parameter $\rho$ by the following function.

$$\rho(c) = \begin{cases} 2\frac{\ln 27/16}{\ln 3/(4c^2)}, & \text{if } c \in (0, 0.131246) \\ \frac{1}{2a_1}\left(-a_2 + \sqrt{a_2^2 - 4a_1 a_3 + 4a_1 c}\right), & \text{if } c \in [0.131246, 0.448994) \\ \frac{1}{2b_3 c}\left(b_1 - b_2 c - \sqrt{(b_1 - b_2 c)^2 - 4b_3 c^2}\right), & \text{if } c \in [0.448994, 0.671256) \\ \frac{1}{2c_3}\left(c_2 - \sqrt{c_2^2 + 4c_3 \ln \frac{3-4c}{4c_1}}\right), & \text{if } c \in [0.671256, 0.75), \end{cases} \quad (26)$$

where $a_i, b_i, c_i$ for $i \in [1, 2, 3]$ are constants as computed by [5] obtained by some polynomial approximations.